\documentclass[10pt,twocolumn,letterpaper]{article}

\usepackage[pagenumbers]{cvpr}      

\usepackage[utf8]{inputenc} 
\usepackage[T1]{fontenc}    
\usepackage{url}            
\usepackage{booktabs}       
\usepackage{amsfonts, amsmath}       
\usepackage{nicefrac}       
\usepackage{microtype}      
\usepackage{xcolor}         
\usepackage{multirow}
\usepackage{graphicx}
\usepackage{subcaption}
\usepackage{dsfont}

\usepackage[pagebackref,breaklinks,colorlinks]{hyperref}

\usepackage[capitalize]{cleveref}
\crefname{section}{Sec.}{Secs.}
\Crefname{section}{Section}{Sections}
\Crefname{table}{Table}{Tables}
\crefname{table}{Tab.}{Tabs.}

\def\paperID{*****}
\def\confName{CVPR}
\def\confYear{2025}

\begin{document}

\title{Exploring Synergistic Ensemble Learning: Uniting CNNs, MLP-Mixers, and Vision Transformers to Enhance Image Classification}

\author{
Mk Bashar\textsuperscript{1} \quad
Ocean Monjur\textsuperscript{2} \quad
Samia Islam\textsuperscript{1} \quad
Mohammad Galib Shams\textsuperscript{3} \quad
Niamul Quader\textsuperscript{4} \\
\textsuperscript{1}Michigan State University \quad
\textsuperscript{2}University of Illinois Urbana-Champaign \\
\textsuperscript{3}Islamic University of Technology \quad
\textsuperscript{4}Motional \\
{\tt\small \{basharmk, islamsa3\}@msu.edu} \quad
{\tt\small omonjur2@illinois.edu} \\
{\tt\small galibshams@iut-dhaka.edu} \quad
{\tt\small niamul.quader@gmail.com}
}

\maketitle

\begin{abstract}
In recent years, Convolutional Neural Networks (CNNs), MLP-mixers, and Vision Transformers have risen to prominence as leading neural architectures in image classification. Prior research has underscored the distinct advantages of each architecture, and there is growing evidence that combining modules from different architectures can boost performance. In this study, we build upon and improve previous work exploring the complementarity between different architectures. Instead of heuristically merging modules from various architectures through trial and error, we preserve the integrity of each architecture and combine them using ensemble techniques. By maintaining the distinctiveness of each architecture, we aim to explore their inherent complementarity more deeply and with implicit isolation. This approach provides a more systematic understanding of their individual strengths.

In addition to uncovering insights into architectural complementarity, we showcase the effectiveness of even basic ensemble methods that combine models from diverse architectures. These methods outperform ensembles comprised of similar architectures. Our straightforward ensemble framework serves as a foundational strategy for blending complementary architectures, offering a solid starting point for further investigations into the unique strengths and synergies among different architectures and their ensembles in image classification. A direct outcome of this work is the creation of an ensemble of classification networks that surpasses the accuracy of the previous state-of-the-art single classification network on ImageNet, setting a new benchmark, all while requiring less overall latency.

\end{abstract}    
\section{Introduction}
\label{sec:intro}
Convolutional neural networks (CNNs), Vision Transformers, and MLP-Mixers are the three fundamental approaches in deep learning for image classification and backbone formulation of other downstream tasks~\cite{alzubaidi2021review,9716741,LIU2022100520}. CNN architectures use convolutional layers that extract local features and hierarchically combine them to recognize complex characteristics of images. They are extremely good at capturing local patterns but are known to struggle when detecting long-range dependencies. On the other hand, Vision Transformers can capture long-range dependencies. The initial transformer architecture was proposed for natural language processing, which requires long-range sequence modeling \cite{vaswani2017attention}. Vision transformers treat images as a sequence of patches and process them using the self-attention mechanism, allowing the model to capture global dependencies \cite{dosovitskiy2021an}. MLP-mixers have a more simplistic design using only multi-layer perceptrons to tackle vision problems. The average inference times are also lower while still being competitive in terms of accuracy compared to Vision transformers and CNNs \cite{tolstikhin2021mlp}.

While the structural build-up of these models is different from one another, we inquire whether various types of models complement each other - i.e. do these models extract fundamentally different features, and are they mutually beneficial when combined? Exploring this question can provide fresh insights into designing more resilient classification network ensembles and backbones for various downstream tasks.

While an increasing number of studies address the aforementioned question \cite{park2021vision} \cite{zhao2021battle} \cite{liu2023learned}, a notable gap remains, in that previous works have not explored ensembles combining these three types of architectures. This is surprising because ensembling is the most systematic and straightforward method for understanding complementarity. Some of the benefits of ensemble learning are improved accuracy, enhanced robustness, versatility, and flexibility \cite{wortsman2022model,benz2021adversarial,croce2023seasoning}. Ensembles have also seen state-of-the-art performance in application domains such as medicine \cite{10190200}, robotics \cite{kasaei2024lifelong}, etc.

Moreover, since understanding complementarity through ensembling does not require combining the architectures of different categories during training, each architecture remains explicitly isolated. This isolation is crucial because it prevents the mixing of features in the learning process, which could differ inherently among the categories, thus providing a purer complementarity study. Therefore, we conduct a large-scale investigation by permuting different types of architecture across the three broad categories.

Our approach focuses on training all models independently and taking the ensemble results of the different types of models. We explore whether combining models with complementary feature extractions indeed improves performance. By exploring the complementarity of classification networks through ensemble learning, we create direct value in formulating better ensembles. The key contributions of this paper are:

\begin{itemize}
\item \textbf{Exploring Complementary Study through Ensembles:} We are pioneers in investigating the complementary characteristics of three fundamental categories of architectures for image classification, maintaining their isolation during training via employing straightforward ensemble techniques.
\item \textbf{Empirical Exploration on Complementary Datasets:} We conduct our complementarity study on CIFAR-10 and ImageNet, which complement each other in terms of size across multiple factors: (1) original image dimensions, (2) number of samples, and (3) number of classes for classification. CIFAR-10 \cite{krizhevsky2009learning} represents a smaller dataset, while ImageNet1k\cite{imagenet15russakovsky} represents a larger and more complex dataset.
\item \textbf{Demonstrating Practical Implications:} We demonstrate that the findings from our complementarity study can lead to more systematic ensembles. Specifically, we build an ensemble that sets a new state-of-the-art on ImageNet, surpassing the closest competitor for accuracy and reduced latency.

\end{itemize}

\section{Related Work}
Prior works have not integrated isolated trained models from different categories of CNNs, MLP-Mixers, and Vision Transformers through ensembles. Neither has there been an empirical exploration of understanding and analyzing the trends of ensembles involving these three types of architecture. However, some prior literature looked into combining modules from these three categories of architecture and tried different approaches to enhance ensembles.

Benz et al. looked into the adversarial robustness between CNNs, Vision Transformers, and MLP-Mixers \cite{benz2021adversarial}. The authors concluded that Vision Transformers is the most robust among the three types of architectures, with CNNs being the worst affected in general by adversarial attacks. They also conclude that CNNs are highly sensitive to high-frequency features, while vision transformers are more sensitive to low-frequency features. Liao et al. state that Transformers are good with long-range dependencies and act as low-pass filters, while convolutions are good with short-range dependencies and act as high-pass filters \cite{liao2023complementary}. They suggest a new network for classifying HSI images by blending convolution methods with a Transformer, adding a Gaussian filter in between to capture mid-frequency data. The authors of \cite{chen2022mobile} present Mobile-Former, a blend of MobileNet and transformer that combines the advantages of local processing and global interaction, efficiently classifying and detecting objects in images better than other models. The paper \cite{zhang2022bending} introduces Trans4PASS for panoramic image segmentation, which incorporates deformable patch embeddings and a deformable MLP module to handle distortions in panoramic images. In \cite{li2022eeg}, the authors introduce TGCNN, a new model combining CNN and Transformer to capture local features and long-range dependencies in EEG signals. 

Zhang et al. introduced a new method using CNN and a unique transformer to accurately count pedestrians from different sources, proving to be the best in cross-modal crowd counting tests \cite{zhang2023cross}. Yuan et al. introduced a new model, CTC-NET, that mixes CNNs and Transformers to better understand medical images by combining their strengths \cite{yuan2023effective}. The authors of the paper \cite{an2023insect} presents a network for insect recognition that blends CNN with attention models, uses a new attention-selection method, and performs better than other models, especially for farming and ecological use.

The combining processes in previous works are based on the heuristic selection of modules that may not generalize to other tasks and require re-conducting the heuristic selection process. Further, training of different modules is not conducted in isolation, which can lead to some modules dominating the results of the overall architecture, which makes the understanding of complementarity among CNNs, Vision transformers, and MLP-mixers difficult. By ensembling models trained in isolation, our approach directly impacts the design of better ensembles moving forward.

\section{Method}
This section outlines the methodologies used to explore the complementary nature of different neural network architectures through ensemble learning, aiming to enhance image classification performance by leveraging the strengths of CNNs, Vision Transformers, and MLP-Mixers.

\subsection{Ensembling Strategy}

The models in our ensembles were all trained in isolation i.e., components of one model had no impact on the other while training. The experiments on the ImageNet dataset were conducted using publicly available weights,  while we trained each model from scratch for the  CIFAR-10 dataset. To ensemble the trained models, the output vectors of each trained model were first stored and subsequently combined using the Softmax operation.

Considering $X^l_{ij}$ as the probability of $i$-th sample of $j$-th class of $X^l$ model, where there $k$ classes and setting the ground truth as $y_i$. The Softmax Score of the $i$-th sample denoted as $S_i$  and the overall Softmax Accuracy score denoted as $SoftAcc$ is calculated as: 

\begin{equation}
  S_i = \arg\max_j(\sum_{l=0}^2 \frac{e^{X^l_{ij}}}{\sum_{j=0}^k e^{X^l_{ij}}})
\end{equation}

\begin{equation}
SoftAcc = \frac{1}{n} \sum_{i=0}^n\mathds{1}(S_i == y_i)
\end{equation}

To quantify the performances of ensembles, accuracy gain denoted as $AccGain$ is used by comparing the highest accuracy among the three ensembled models and the Softmax accuracy denoted as $SoftAcc$.

\begin{equation}
\begin{aligned}
    AccGain = SoftAcc - \max( & Accmodel_1, Accmodel_2, \\
     & Accmodel_3)
\end{aligned}
\end{equation}

\subsection{Quantitative and Qualitative Analysis}
\subsubsection{Partial Correlation Analysis}
We conducted a partial correlation analysis of samples where at least one of the three models in our ensemble made a mistake. By examining the partial correlations between the model outputs, we can identify whether different models are learning distinct and complementary features. This reveals even if models making mistakes on different samples can be combined together to get better classification accuracy. Understanding the trend of output correlation and ensembled accuracy gain can highlight how ensembles can benefit from the complementary strengths of individual models.

To calculate the partial correlation among three model predictions, we designated each of the model predictions as variables $X, Y, Z$. The correlation between two variables for all the combinations $X$ and $Y$, $X$ and $Z$, $Y$ and $Z$ is denoted as $r_{XY}$, $r_{XZ}$ and $r_{YZ}$ respectably. From this, the multiple correlation coefficient is defined as:

\begin{equation}
  R_{Z,XY} = \sqrt{\frac{r_{XZ}^2 + r_{YZ}^2 - 2 r_{XY} r_{YZ} r_{XZ}}{1-r_{XY}^2}}
\end{equation}

In this formula, $X$ and $Y$ are independent variables, while $Z$ is the dependent variable. For an unbiased insight, an adjusted version was taken using the formula from \cite{schmuller2021statistical}:

\begin{equation}
  R_{adj}^2 = 1 - \frac{(1-R_{Z,XY}^2)(n-1)}{n-k-1}
\end{equation}

Here, $n$ represents the number of samples, and $k$ is the count of the independent variables. This adjusted $R_{adj}$ provides a nuanced view of how the predictions from different models correlate.

\subsubsection{Gradient Weighted-Class Activation Maps}
To visually explore the complementary characteristics of CNNs, Vision Transformers, and MLP-Mixers, the last layers of these models are selected to visualize the GradCAMs. We want to visually inspect the models which together, have the highest accuracy gains among all the ensembles. If the models are truly complementary and if this characteristic plays a role in ensembled performance gains, ensembles with high accuracy gains should show differences in visual results. Additionally, we aim to analyze whether the visual differences in GradCAMs correlate with the quantitative performance improvements observed in the ensemble, thereby reinforcing the hypothesis that the diversity in model architectures contributes significantly to enhanced image classification accuracy.

\subsubsection{Average Relative Log amplitude of Fourier transformed feature maps}
\label{Fourier}
We examine the average frequency domain feature representation of each type of model by averaging the frequency domain feature representation for the last layer of CNNs, Vision Transformers, and MLP-Mixers. While the GradCAM outputs can be inconclusive, by analyzing the average frequency domain we can get a quantifiable metric to analyze and compare complementarity among different model architectures. We performed frequency domain analysis on all 10,000 images from the CIFAR-10 test set, and for ImageNet, we randomly selected 10,000 images for all of the models.

\subsection{Experimental Setup}
\label{sec:ExSetup}
All the required training and inference were conducted on a single NVIDIA RTX 3080 with 12GB of VRAM. For a fair comparison of each model, we tried to limit the final accuracy of all our models to around 80\% for ImageNet by selecting different variations of popular models. For CIFAR-10, the final accuracy was limited to 85\% by incorporating early stopping. The idea behind this approach stems from the fact that we do not want any individual model to dominate the ensembling results. Allowing the participating models in an ensemble to start from an unbiased position, ensuring equal potential for accuracy improvements within the ensemble. The ensembles are constructed with every possible permutation of 3 models.

\begin{table*}
\centering
\resizebox{\linewidth}{!}{%
\begin{tabular}{ccccccccc}
\toprule
Category & Model 1 & Acc($\%$)  & Model 2  & Acc($\%$) & Model 3 & Acc($\%$) &  SoftAcc($\%$)& AccGain($\%$)\\
\hline
\multirow{ 5}{*}{ImageNet} & SENet    & 81.37    & Swin-T    & 80.55 & ViT  & 81.40 & 83.85 & 2.44 \\
& ConvNeXt    & 79.42   & DPN & 79.68 & DeiT & 79.71 & 82.15 & 2.43 \\
 & EfficientNet & 81.02   & SENet   & 81.37  & ViT  & 81.40 & 83.79 & 2.39\\
 & CrossViT & 80.86   & SENet    & 81.37  & ViT  & 81.40 & 83.78& 2.37\\
 & DPN & 79.68   & SENet    & 81.37  & ViT  & 81.40 & 83.76 & 2.35\\
\hline
\multirow{ 5}{*}{CIFAR-10} & GoogleNet    & 85.46   & MobileNet    & 85.15 & VisFormer  & 85.10 & 91.46 & 6.22 \\
& MobileNet    & 85.15    & ResNet & 85.01 & VisFormer & 85.10 & 91.23 & 6.14 \\
& DPN & 85.40   & ResNet    & 85.01  & VisFormer  & 85.10 & 91.28& 6.11\\
& ResNet    & 85.01   &SeNet & 85.61 & VisFormer & 85.10 & 91.29  & 6.05 \\
& MobileNet    & 85.15    & SqueezeNet & 86.50 & VisFormer & 85.10 & 91.60 & 6.02 \\
\bottomrule
\end{tabular}
}
\caption{Top five ensembles in terms of fair gain in accuracies, where the ensemble is based on adding softmax scores.}
\label{tab:1}
\end{table*}

\subsubsection{ImageNet1k}
Experiments on the ImageNet1k dataset\cite{imagenet15russakovsky} were conducted via running inference on the validation set using pre-trained models from \cite{rw2019timm} rather than training them from scratch. Our experiments included 7 CNN, 6 Transformer-based architectures, and
2 MLP-Mixers. These 15 architectures gave us a total of 455 ensembled combinations. We selected these models based on their high impact, widespread availability, and popularity in usage.

\textbf{Selected Architectures:} The following 7 models were selected as the CNNs for our experiments.
ResNet \cite{7780459}, ResNeXt \cite{8100117}, RepVGG \cite{ding2021repvgg}, SENet \cite{hu2018squeeze}, ConvNeXt \cite{liu2022convnet}, EfficientNet \cite{tan2019efficientnet}, and DPN \cite{chen2017dual}. For Vision Transfomers we used, Swin \cite{liu2021swin}, Visformer \cite{chen2021visformer}, VIT \cite{dosovitskiy2021an}, PVT \cite{wang2021pyramid}, DeIT \cite{touvron2021training}, and CrossVIT \cite{chen2021crossvit}. The 2 MLP-Mixers for our experiments are  ResMLP \cite{touvron2022resmlp} , gMLP \cite{liu2021pay}.
\subsubsection{CIFAR-10}
For CIFAR-10 \cite{krizhevsky2009learning}, all of the models were trained from scratch. Our experiments on CIFAR-10 had 9 CNNs, 4 MLP-Mixer, and 4 Transformer-based architectures. With a total of 680 ensembled combinations.

\textbf{Selected Architectures:} ResNet \cite{7780459}, ResNeXt \cite{8100117}, VGG \cite{simonyan2014very}, SENet \cite{hu2018squeeze}, SqueezeNet \cite{iandola2016squeezenet}, DPN \cite{chen2017dual}, EfficientNet \cite{tan2019efficientnet}, GoogleNet \cite{szegedy2015going}, MobileNet \cite{howard2017mobilenets} were selected as the CNNs. 
For the Vision Transfomers we used Swin \cite{liu2021swin}, Visformer \cite{chen2021visformer}, MaxViT \cite{tu2022maxvit}, and PVT \cite{wang2021pyramid}.
The 4 MLP-Mixers for our experiments are  SparseMLP \cite{tang2022sparse}, WaveMLP \cite{tang2022image}, SwinMLP \cite{liu2021swin}, MorphMLP \cite{zhang2022morphmlp}.

\section{Results}
Our approach to determining complementarity and formulating a systematic strategy for ensemble construction involves a detailed analysis of raw ensembled accuracy results and a thorough exploration of feature-level disparities inherent in each architectural type. These findings shed light on the synergistic effects between different architectural types, particularly the notable performance boosts achieved through combinations of Vision Transformers and CNNs.

\subsection{Complementarity analysis via accuracy gains}
\label{ensemble}
\textbf{Analysis of Top 5 Ensembled Accuracies:}
Table \ref{tab:1} shows the top 5 ensembles by accuracy gain for both ImageNet and CIFAR-10. For ImageNet, the top 5 ensembles have gains of around 2.4\%, with the combinations containing a mix of transformers and CNNs. For CIFAR-10, the top 5 ensembles have large gains in performance, with accuracy gains of over 6\%. Interestingly, the top ensembles have exactly two CNNs and one Vision Transformer. For both the datasets, the Combination of CNNs and Transformers gives the highest accuracy gain, implying that the combination of Vision Transformers and CNNs boosts ensemble performance compared to other combinations.

\begin{table*}
\centering
\resizebox{\linewidth}{!}{%
\begin{tabular}{ccccccccc}
\toprule
Category & Model 1 & Acc($\%$)  & Model 2  & Acc($\%$) & Model 3 & Acc($\%$) &  SoftAcc($\%$)& AccGain($\%$)\\
\hline
\multirow{ 3}{*}{CMT} & SeNet    & 81.37    & ViT    & 81.40 & ResMLP  & 80.54 & 83.62 & 2.16 \\
& DPN  & 79.68    & Swin & 80.55 & ResMLP & 80.54 & 82.63 & 2.11 \\
 & EfficientNet & 81.02   & Swin    & 80.55  & ResMLP & 80.54 & 83.05 & 2.02\\
\hline
CNN & DPN    & 79.68    & ResNet    & 80.08 & RepVGG  & 80.21 & 82.45 & 2.23 \\
Transformer & PVT    & 81.68    & ViT & 81.40 & VisFormer & 81.82 & 83.80 & 1.97 \\
MLP & ResMLP & 80.54   & gMLP    & 77.204  & ---  & --- & 80.16& -.38\\
\bottomrule
\end{tabular}
}
\caption{ImageNet Results on the top ensembles from completely mixed categories (i.e. CNNs, MLP-Mixers and Transformers - abbreviated as CMT) and completely homogeneous categories (i.e. ensemble of CNNs only or MLP-Mixers only or Transformers only). Mixed models have a tendency to provide much better gains in accuracy.}
\label{t2}
\end{table*}

\begin{table*}
\centering
\resizebox{\linewidth}{!}{%
\begin{tabular}{ccccccccc}
\toprule
Category & Model 1 & Acc($\%$)  & Model 2  & Acc($\%$) & Model 3 & Acc($\%$) &  SoftAcc($\%$)& AccGain($\%$)\\
\hline
\multirow{ 3}{*}{CMT} & SqueezeNet    & 86.50    & SwinMLP    & 85.57 & VisFormer  & 85.10 & 91.45 & 5.77 \\
& GoogleNet    & 85.46    & SwinMLP & 85.57 & VisFormer & 85.10 & 91.06 & 5.68 \\
 & ResNet & 85.01   & SwinMLP    & 85.57  & VisFormer  & 85.68 & 90.86& 5.63\\
\hline
CNN & GoogleNet    & 85.46    & ResNet    & 85.01 & ResNeXt  & 85.20 & 90.89 & 5.67 \\
Transformer & MaxVit    & 85.90    & Swin & 86.52 & VisFormer & 85.10 & 90.23 & 4.39 \\
MLP & SparseMLP & 85.95   & SwinMLP    & 85.57  & WaveMLP  & 85.68 & 90.17& 3.44\\
\bottomrule
\end{tabular}
}
\caption{CIFAR-10 results on the top ensembles from completely mixed categories (i.e. CNNs, MLP-Mixers and Transformers - abbreviated as CMT) and completely homogeneous categories (i.e. ensemble of CNNs only or MLP-Mixers only or Transformers only). Mixed models have a tendency to provide much better gains in accuracy.}
\label{t3}
\end{table*}

\textbf{Analysis of Ensembles containing one from each type of Architecture:}
The CMT(CNN, MLP-Mixer, and Vision Transformer) category of Tables \ref{t2} and \ref{t3} shows the top 3 performing ensembles containing one of each type of architecture from ImageNet and CIFAR-10. For ImageNet, the accuracy gains from using all three models show a decrease from the gains seen using a combination of Vision transformers and CNNs. Similarly, for CIFAR-10, a similar drop is noticeable in accuracy gains across all other types of combinations except CNNs and Vision Transformers. This leads us to believe that the introduction of MLP-Mixers in the ensembles does not help the overall performance compared to the sole combinations of CNNs and Vision Transformers.

The performance drop when incorporating MLP-Mixers into the ensembles for both ImageNet and CIFAR-10 suggests that MLP-Mixers might not be complementary to CNNs and Vision Transformers. The addition of MLP-Mixers may introduce redundancy or disrupt the synergy between CNNs and Vision Transformers, thereby diluting the performance gains that are achieved with just these two architectures.

\textbf{Analysis of Ensembles with only one type of Architectures:}
Results from Tables \ref{t2} and \ref{t3} strengthen our hypothesis that complementarity does lead to better ensemble performance, particularly combinations of CNNs and Vision Transformers. When looking at the top-performing ensembles containing the same type of architecture, we see a decrease in accuracy gain compared to the mixed ensemble of CNNs and Transformers in Table 1. The combination of 3 CNNs gives a relatively higher accuracy gain for both ImageNet and CIFAR-10, compared to the combinations containing only Vision Transformers or MLP-Mixers.

\subsection{Feature-level complementarity analysis}

\begin{figure*}
  \centering
  \begin{subfigure}[t]{0.49\textwidth}
  \centering
   \includegraphics[width=\linewidth, trim={0 0 1.5cm 0},clip]{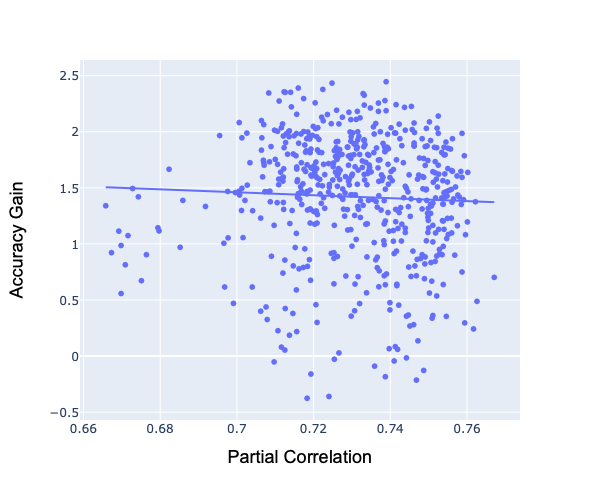}
      \caption{Imagenet1k}
  \end{subfigure}
  ~
  \begin{subfigure}[t]{0.49\textwidth}
      \centering
      \includegraphics[width=\linewidth, trim={0 0 1.5cm 0},clip]{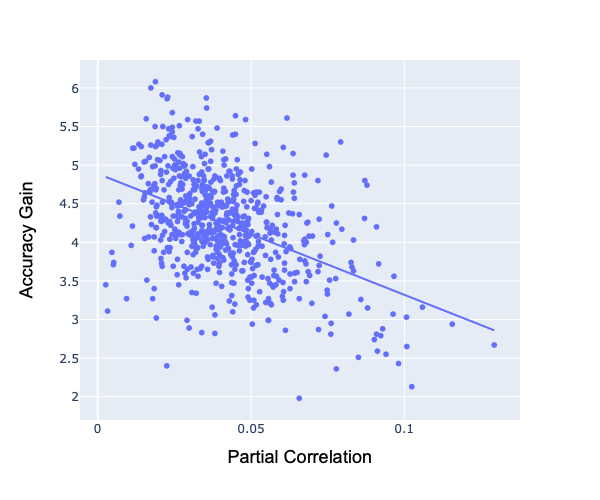}
      \caption{Cifar-10}
  \end{subfigure}
   \caption{Accuracy Gain vs Partial Correlation for all ensembles. There is a trend of decreased accuracy gain with increased partial correlation in both datasets. The trend is more visible in the CIFAR-10 dataset}
   \label{fig:one}
\end{figure*}

\textbf{Partial Correlation Analysis:}
The results from Figure \ref{fig:one} reveals the following trend: as the partial correlation decreases, the accuracy gain of the ensemble increases. This trend is observed in both datasets, suggesting that models with lower partial correlation, which indicates they are learning more complementary outputs, contribute to better ensemble performance.

This trend is much steeper for CIFAR-10 compared to ImageNet. The explanation for this can be drawn from the fact that for CIFAR-10, the accuracy gains are more evident as most ensembles grow over 4-5\%. For ImageNet, the trend is less clear but still observable.

These findings underscore the importance of combining models that learn diverse features to enhance the performance of ensemble methods, particularly for datasets where such complementarity can yield significant gains. This analysis highlights that maximizing the diversity of model outputs, as indicated by lower partial correlation, can be crucial for optimizing the effectiveness of ensemble strategies.

\begin{figure*}
    \centering
    \resizebox{\linewidth}{!}{%
\begin{tabular}{@{\hspace{0.2em}}c@{\hspace{0.2em}}@{\hspace{0.2em}}c@{\hspace{0.2em}}@{\hspace{0.2em}}c@{\hspace{0.2em}}@{\hspace{0.2em}}c@{\hspace{0.2em}}}
Original Image & SeNet & Swin & ViT\\
\includegraphics[trim={1cm 1cm 1cm 1cm},clip, width=.23\textwidth]{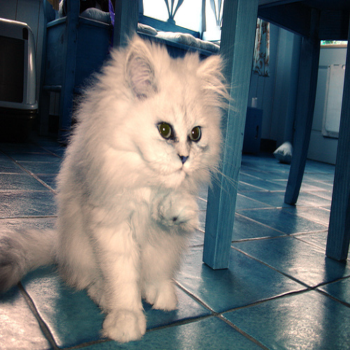} & \includegraphics[trim={1cm 1cm 1cm 1cm},clip, width=.23\textwidth]{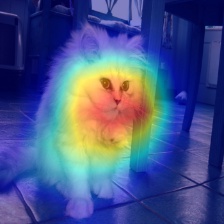}
& \includegraphics[trim={1cm 1cm 1cm 1cm},clip, width=.23\textwidth]{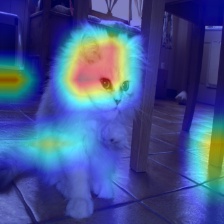}& \includegraphics[trim={1cm 1cm 1cm 1cm},clip, width=.23\textwidth]{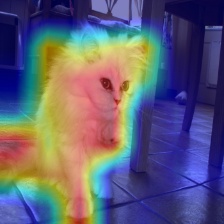} \\
\includegraphics[trim={1cm 1cm 1cm 1cm},clip, width=.23\textwidth]{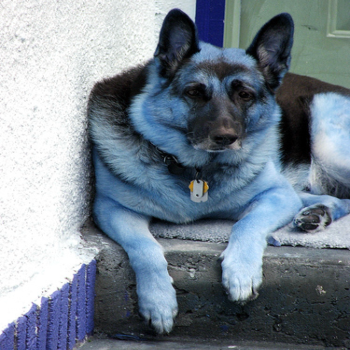} & \includegraphics[trim={1cm 1cm 1cm 1cm},clip, width=.23\textwidth]{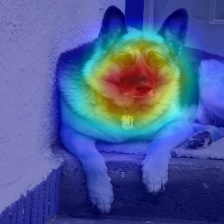}
& \includegraphics[trim={1cm 1cm 1cm 1cm},clip, width=.23\textwidth]{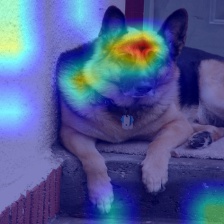} & \includegraphics[trim={1cm 1cm 1cm 1cm},clip, width=.23\textwidth]{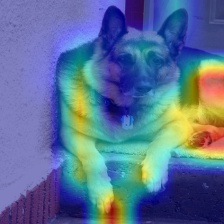}
\end{tabular}
}
\caption{Gradient Activation Maps of two sample images from Imagenet on the best ensembled combination (SeNet, Swin, ViT). All three models visually show distinct characteristics implying their complementarity. This aligns with our average frequency analysis across all models \ref{fig:two} }
\label{i1}
\end{figure*}

\begin{figure*}
  \centering
  \begin{subfigure}[t]{0.49\textwidth}
  \centering
  \includegraphics[width=\linewidth, trim={0 0 1.5cm 0},clip]{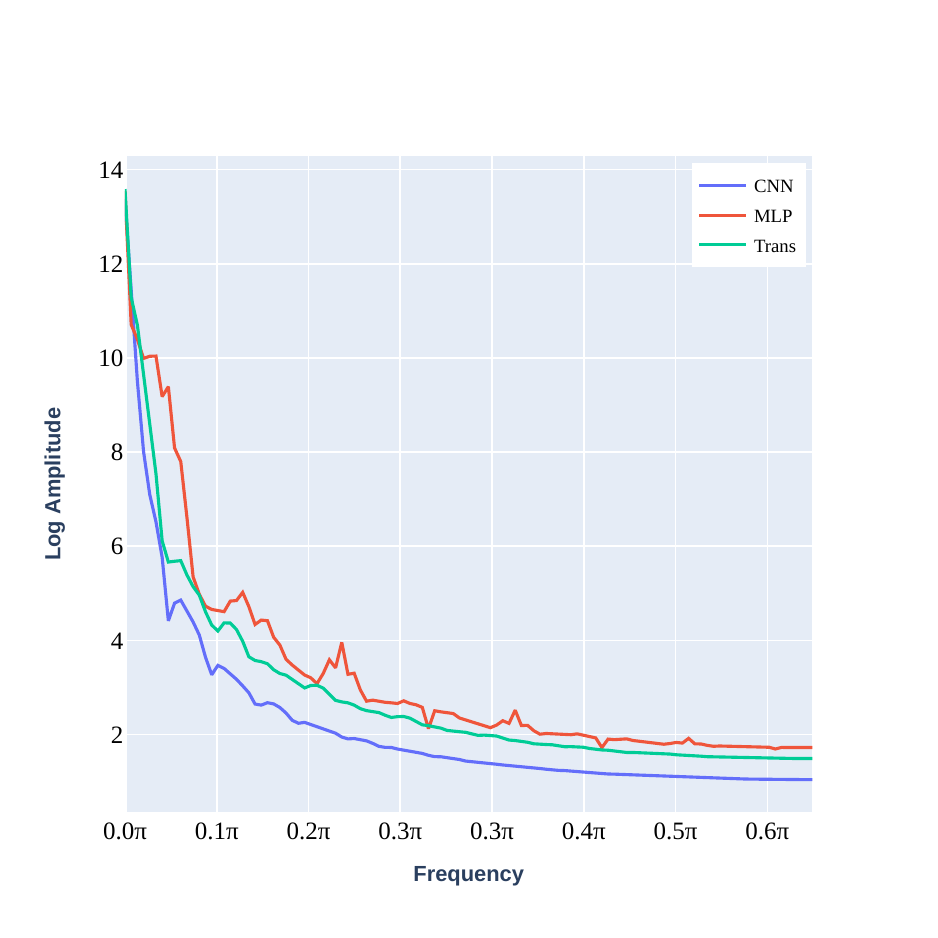}
    \caption{Imagenet}      
  \end{subfigure}
  ~
  \begin{subfigure}[t]{0.49\textwidth}
      \centering
      \includegraphics[width=\linewidth, trim={0 0 1.5cm 0},clip]{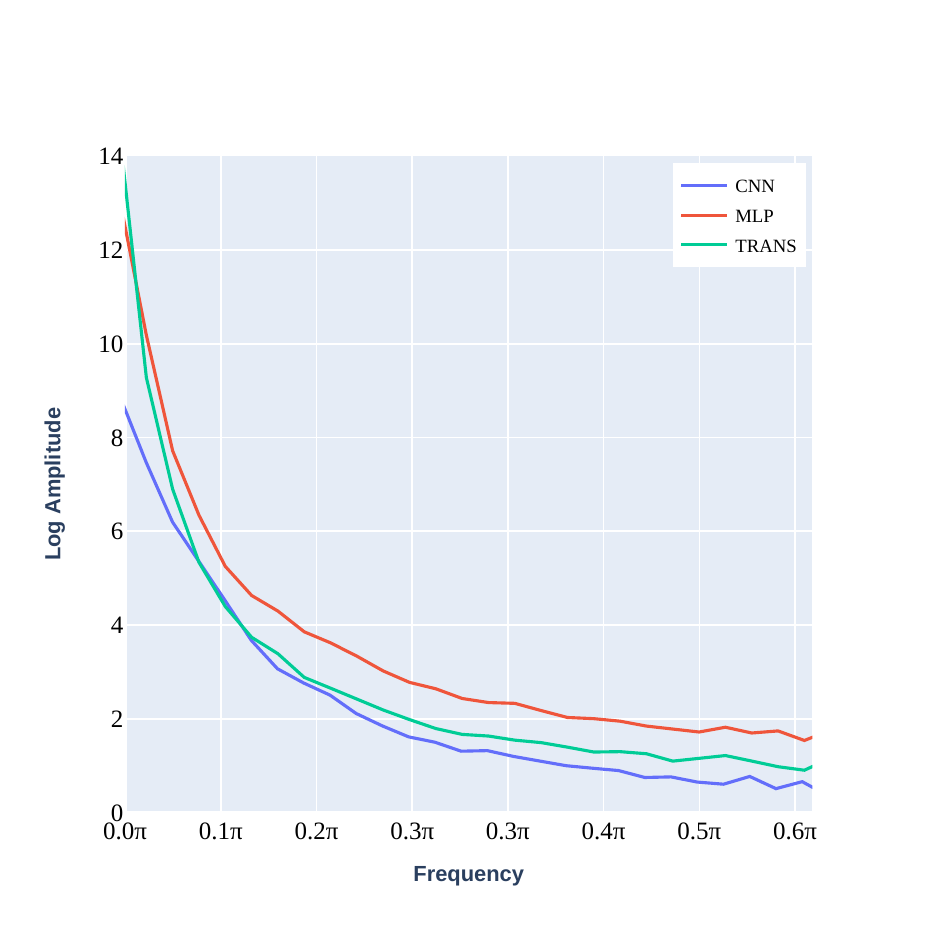}
      \caption{Cifar-10}
  \end{subfigure}
   \caption{Log Amplitude vs. Frequency graph of intermediate features in CNNs, MLP, and Transformers. The frequency characteristics of CNNs and MLP-Mixers are more similar or correlated than that of CNNs and Transformers.}
   \label{fig:two}
\end{figure*}

\begin{table*}

\centering
\resizebox{\linewidth}{!}{%
\begin{tabular}{cccccccccc}
\toprule
Category & Model 1 & Acc($\%$)  & Model 2  & Acc($\%$) & Model 3 & Acc($\%$) &  SoftAcc($\%$)& AccGain($\%$) & Inf. time(s)\\
\hline
\multirow{ 1}{*}{Top-3} & EVA-02    & 89.56   & EVA-02v2   & 89.24 & EVA-Giant  & 89.61 & 89.942 & .332 & .409 \\
\multirow{ 1}{*}{Top-3 ViTs} & BeiTv2    & 87.93   & EVA-Giant    & 89.61 & ViT  & 88.48 & 89.66 & .05 & .407 \\
\multirow{ 1}{*}{CTT} & ConvNeXtV2    & 88.84   & EVA-02  & 89.56 & EVA-02v2  & 89.24 & \textbf{90.03} & \textbf{.470} & \textbf{.222}\\
\bottomrule
\end{tabular}
}
\caption{Comparison of Ensembling using CTT vs ensembling the best performing models.CTT outperforms combinations even if the selected models are better in other combinations}
\label{tab:4}
\end{table*}

\begin{figure*}
  \centering
  \begin{subfigure}[t]{0.49\textwidth}
  \centering
   \includegraphics[width=\linewidth, trim={0 0 1.5cm 0},clip]{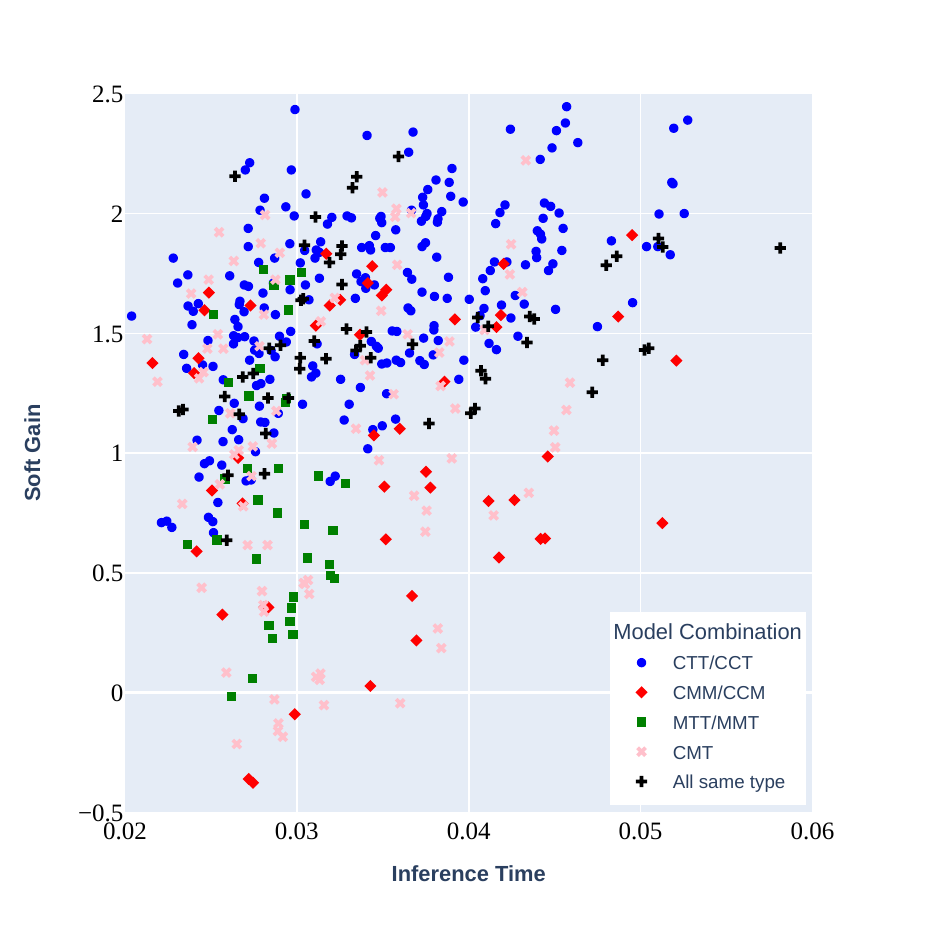}
      \caption{Imagenet1k}
  \end{subfigure}
  ~
  \begin{subfigure}[t]{0.49\textwidth}
      \centering
      \includegraphics[width=\linewidth, trim={0 0 1.5cm 0},clip]{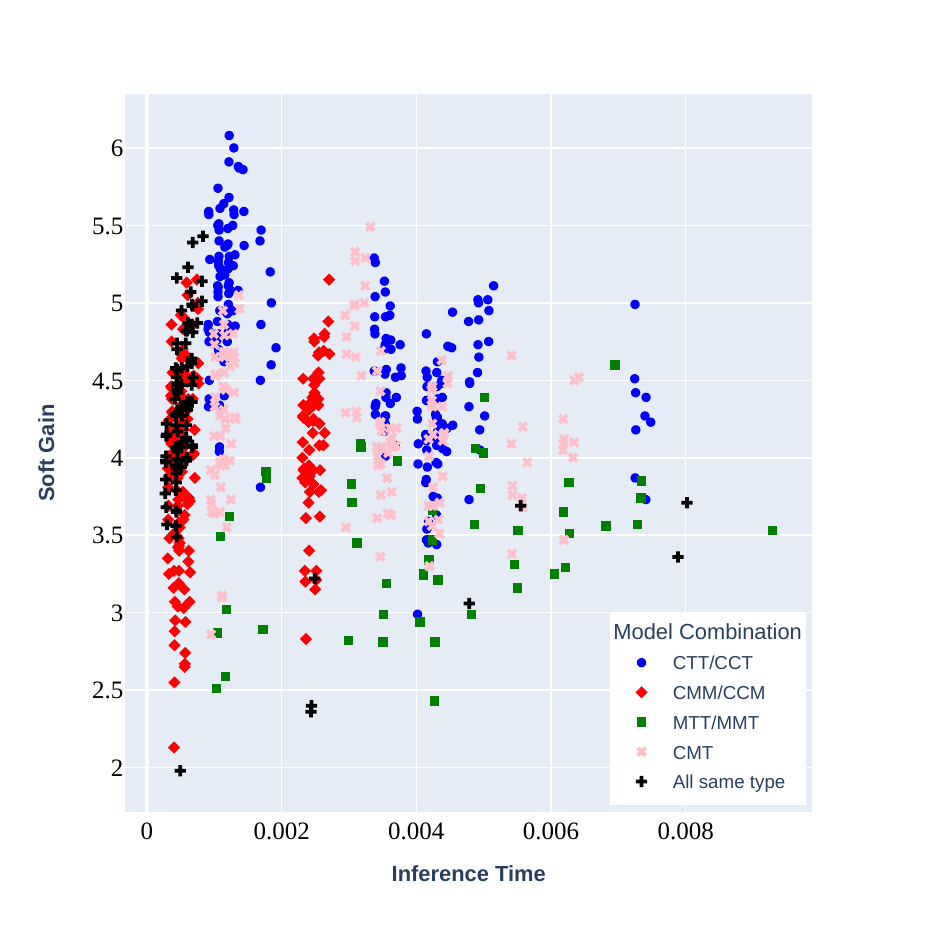}
      \caption{CIFAR-10}
  \end{subfigure}
   \caption{Softmax ensemble accuracy gain vs inference time for different types of ensembles. A common trend that can be observed is that combinations mixing CNNs and Transformers tend to outperform all other types of combinations among different inference times}
   \label{fig:onecol}
\end{figure*}

\textbf{Gradient Activation Map:}
For a more detailed analysis of the outcomes, we also examined the inter-model prediction correlations, focusing specifically on the synergistic interplay between three models whose ensemble yielded the highest accuracy gain for ImageNet. 

The ensemble of SeNet, Swin, and ViT had the highest accuracy gain on the ImageNet experiments. We arbitrarily chose two images from the ImageNet validation set to compare and contrast how these models behave. The analysis from Figure \ref{i1} indicates that SeNet and Swin consistently focus on a small portion of the object. The activation area of the Swin model is smaller than SeNet. ViT, distinctively, often triggers activation over the total object of interest. Similar findings in other samples further reinforce our quantitative observation that CNNs and Vision Transformers exhibit unique perspectives and tend to complement each other within ensembles.

\textbf{Frequency Domain Analysis:}
Now that we have analyzed complementarity at the output classification layer via correlation analysis, we proceed to analyze feature-level complementarity. From Figure \ref{fig:two} it is evident that CNNs have the lowest amplitude across frequencies compared to Vision Transformers and MLP-Mixers. We see a clear distinction in the frequency domain distribution and amplitude across frequencies between Vision transformers and CNNs. For MLP-Mixers, the frequency magnitude is higher than both Vision Transformers and CNNs, but the frequency domain distribution follows the same pattern as CNNs, implying the similarity between CNNs and MLP-Mixers. These findings reinforce our previous result that MLP-Mixers might not be as complementary as CNNs are with Vision Transformers.

\subsection{Benchmarking Ensembles}
Through an extensive empirical investigation encompassing all potential combinations among the chosen architectural types, a discernible pattern can be observed when comparing accuracy gains with inference time shown in Figure \ref{fig:onecol}. Across all inference time intervals, combinations incorporating CNNs and Vision Transformers demonstrate the most substantial improvements in accuracy for both CIFAR-10 and ImageNet. Our discoveries remain consistent across the smaller CIFAR-10 and the larger ImageNet dataset, reinforcing the complementarity of CNNs and Vision Transformers and evidence on generalizability across different datasets.

Based on the above-identified paradigm that ensembles of CNNs and Vision Transformers provide positive results for increased performance and reduced latency, we combine two Vision Transfomers (EVA-02, EVA-02v2) and one CNN (ConVNeXtv2) from the top-performing models on timm \cite{rw2019timm} - results ( Table \ref{tab:4}) show that the ensemble performs better (increased performance, lower latency) compared to ensembles of same types of networks. We compared this ensemble against ensembles formed by the top-3 models available in the timm \cite{rw2019timm} library. As the top 3 models are all variants of EVA, we also compared them against the top 3 ViTs of distinct types. The findings presented in Table \ref{tab:4} demonstrate that employing a combination of CNNs and Vision Transformers not only yields higher accuracy gains but also results in lower inference times compared to other types of combinations. This reinforces our hypothesis that complementarity among architectures yields overall better performance. Furthermore, the ensemble performs better than the best-performing individual model EVA-Giant (Accuracy: 89.61\% and latency: 0.28sec).

\section{Limitations}

The experiments in this paper were only conducted for image classifications.The limitation of this paper lies in analyzing MLP-Mixers for the ImageNet dataset. We used only two MLP-Mixers due to the limited availability of pre-trained MLP-Mixers. However, the results were still consistent with that of CIFAR-10, which leads us to believe our experimental results will still hold when adding more MLP-Mixers to the ensembles.

\section{Conclusion}
In this work, we conduct an empirical exploration into the complementarity of CNNs, Vision Transformers, and MLP-Mixers. We began with the hypothesis that complementarity will lead to better ensemble performance. The results show that mixing different types of architecture provides a larger performance gain than just mixing the same type of architecture. Furthermore, we found that CNNs and Vision Transformers complement each other the most when considering the ensembled accuracy gain for all ranges of inference times. By investigating both CIFAR-10 and ImageNet datasets, this study encompasses both smaller and larger dataset sizes, thereby enhancing the validity of the findings for generalization to other datasets. Additionally, we demonstrate that our findings can be extended to the discovery of new state-of-the-art ensembles, surpassing those formed by arbitrarily mixing models. It lays down important groundwork and motivates future efforts, including but not limited to identifying effective and systematic image classification ensembles, creating robust backbone formulations for downstream tasks like detection and segmentation, and exploring the explainable complementarity of different architectures.

\label{limit}
{\small
\bibliographystyle{ieee_fullname}
\bibliography{main}
}

\end{document}